\documentclass{article}

\usepackage[nonatbib, final]{neurips_2024}
\usepackage{orcidlink}

\usepackage[utf8]{inputenc} 
\usepackage[T1]{fontenc}    
\usepackage{hyperref}       
\usepackage{url}            
\usepackage{booktabs}       
\usepackage{amsfonts}       
\usepackage{nicefrac}       
\usepackage{microtype}      
\usepackage{xcolor}         
\usepackage{tikz}           
\usepackage{appendix}       
\usepackage[sortcites, backend=biber,sorting=none]{biblatex}
\usepackage{multirow}

\title{Explicitly Modeling Pre-Cortical Vision with a Neuro-Inspired Front-End Improves CNN Robustness}

\author{%
  Lucas Piper\orcidlink{0009-0000-5963-1163}$^{1}$, Arlindo L. Oliveira\orcidlink{0000-0001-8638-5594}$^{1,2,}$, Tiago Marques\orcidlink{0000-0002-8973-0549}$^{3}$ \\
  \\
  $^1$IST Técnico Lisboa, Universidade de Lisboa, Portugal \\
  $^2$INESC-ID Lisboa, Portugal \\
  $^3$Breast Unit, Champalimaud Clinical Center, Champalimaud Foundation, Lisboa, Portugal \\
  \texttt{lucaspiper99@tecnico.ulisboa.pt} \\ \texttt{tiago.marques@research.fchampalimaud.org} \\
}

\begin{document}

\maketitle

\begin{abstract}
    While convolutional neural networks (CNNs) excel at clean image classification, they struggle to classify images corrupted with different common corruptions, limiting their real-world applicability. Recent work has shown that incorporating a CNN front-end block that simulates some features of the primate primary visual cortex (V1) can improve overall model robustness. Here, we expand on this approach by introducing two novel biologically-inspired CNN model families that incorporate a new front-end block designed to simulate pre-cortical visual processing. RetinaNet, a hybrid architecture containing the novel front-end followed by a standard CNN back-end, shows a relative robustness improvement of 12.3\% when compared to the standard model; and EVNet, which further adds a V1 block after the pre-cortical front-end, shows a relative gain of 18.5\%. The improvement in robustness was observed for all the different corruption categories, though accompanied by a small decrease in clean image accuracy, and generalized to a different back-end architecture. These findings show that simulating multiple stages of early visual processing in CNN early layers provides cumulative benefits for model robustness.
\end{abstract}

\section{Introduction}

Convolutional neural networks (CNNs) excel in object recognition \cite{NIPS2012_c399862d, simonyan2015deep, szegedy2014going, he2015deep, huang2018densely, tan2020efficientnet, cai2023reversible} but struggle with generalizing to different datasets, limiting their real-world applicability \cite{recht2019imagenet, engstrom2019exploring, alcorn2019strike, tian2018deeptest, hendrycks2019benchmarking}. This gap in robustness highlights differences between CNNs and human vision, including how they process visual information \cite{Lindsay_2021}, their susceptibility to errors \cite{geirhos2021partial, geirhos2020accuracy} and to adversarial attacks \cite{szegedy2014intriguing, moosavidezfooli2017universal, kurakin2017adversarial}.

Recent research has focused on enhancing CNN robustness by drawing insights from neuroscience \cite{FEDERER2020103, 23, MALHOTRA202057, Dapello2020.06.16.154542, safarani2021robust}. Notably, VOneNets \cite{Dapello2020.06.16.154542}, a family of CNNs constructed by introducing a biologically-inspired front-end block followed by a trainable CNN architecture, have shown improved robustness against adversarial attacks and common image corruptions. The front-end block, the VOneBlock, simulates the primate primary visual cortex (V1) by incorporating a fixed-weight, data-constrained Gabor Filter Bank (GFB). However, this approach overlooks explicit modeling of prior visual processing stages, raising the question: \textbf{can the explicit modeling of the retina and the lateral geniculate nucleus (LGN) further improve model robustness?}

In this work, we make the following key contributions:
\begin{itemize}
    \item  We introduce a novel fixed-weight CNN front-end block called the RetinaBlock, designed to simulate the retina and the LGN, operating as a multi-stage cascading linear-nonlinear model parameterized by neurophysiological studies.
    \item We introduce two novel CNN families: RetinaNets and EVNets (Early Vision Networks). RetinaNets integrate the RetinaBlock followed by a standard CNN back-end architecture, while EVNets couple the RetinaBlock with the VOneBlock before the back-end.
    \item We show that both new CNN families improve robustness to common corruptions when compared to the base model and that the gains introduced by the RetinaBlock stack with those due to the VOneBlock.
    \item We verify that these robustness gains generalize to other model architectures, by testing both a ResNet18 and a VGG16 variant of each family.
\end{itemize}

\subsection{Related Work}

\paragraph{Retina modeling.} The spatial summation over the receptive fields (RFs) of retinal ganglion cells in primates was first described by the Difference-of-Gaussian (DoG) model \cite{kuffler1953, Rodieck1965-yv}. This model was later expanded to account for extra-classical RF effects such as contrast gain control \cite{Shapley1978-dx, Solomon2006-zt}. Divisive normalization mechanisms \cite{Carandini2011-xb, Bonin2005-cu} along with linear-nonlinear-linear-nonlinear (LNLN) frameworks \cite{MANTE2008625} have further improved retinal response prediction by describing the interaction between different visual processing stages. More recently, CNNs have outperformed prior models in predicting retinal responses to visual stimuli \cite{McIntosh2016}.

\paragraph{Common corruptions.} To assess out-of-domain generalization, datasets have been developed to incorporate common corruptions and different rendition styles \cite{hendrycks2021natural, xie2020adversarial}. As for improving model generalization ability, data augmentation techniques have been a popular approach \cite{hendrycks2020augmix, hendrycks2021faces, 2001.06057}, with recent work demonstrating improved performance by leveraging compositions of augmentation operations \cite{hendrycks2020augmix} and training data from an image-to-image network \cite{hendrycks2021faces}. Other generalization approaches include knowledge distillation \cite{hong2021studentteacher} and masking activations to balance learning between domain-invariant and domain-specific features \cite{chattopadhyay2020learning}.


\paragraph{Neuro-inspired models.}  Convolution layers that simulate RFs of early vision neurons have been shown to improve model robustness \cite{EVANS202296, 10.1007/978-3-031-44204-9_33}. Modeling V1 in front of CNNs improves white-box adversarial robustness and performance for common corruptions \cite{Dapello2020.06.16.154542}. Additionally, incorporating a divisive normalization layer produces further gains in robustness besides a higher alignment with V1 responses \cite{cirincione2022implementing}. Other neuro-inspired strategies involve summing activation maps from filters with opposite polarity to simulate the push-pull inhibition pattern found in V1 \cite{pushpull2024, 10.1007/s00521-020-04751-8} and using a multi-task training strategy to perform image classification while predicting neural patterns \cite{FEDERER2020103, safarani2021robust}

\section{Methods} \label{sec:methods}

Here, we introduce two novel CNN families called RetinaNets and EVNets, comprising the RetinaBlock as illustrated in Figure \ref{fig:1}, following a similar approach as used in VOneNets (see Supplementary Material Section \ref{sec:vonenets}). The RetinaBlock models foveal visual processing in the retina and LGN, whereas the assembly of the RetinaBlock and VOneBlock models processing up to V1. RetinaNets and EVNets operate as firing-rate models, eschewing explicit temporal dynamics to focus on spatial processing. We used a simplified method where cone responses were approximated with RGB values, without scaling to match model activations with empirical spike train frequencies.  All models were set to a 2-degree field-of-view, consistent with previous adaptations of the VOneBlock to the Tiny ImageNet dataset (see Supplementary Material Section \ref{sec:datasets}) \cite{baidya2021combining, barbulescu2023matchingneuronalrepresentationsv1}.


The RetinaBlock  simulates spatial summation over the extra-classical RF of midget and parasol retinal ganglion cells, processing them as separate parallel pathways. The midget cell pathway consists of a light-adaptation layer and a DoG convolutional layer, whereas parasol cells also have a contrast-normalization layer to reflect the contrast gain control observed in these cells \cite{Solomon2006-zt, RaghavanENEURO.0515-22.2023, Lee1994-co}.

\paragraph{Push-pull pattern.} The push-pull pattern characterized by the interaction of on- and off-center cells \cite{doi:10.1126/science.8191289, Hirsch1998-lq} can be modeled by incorporating opposite-polarity filters, rectifying activations, and subtracting on-center from off-center activations. Due to the symmetry of the cells, the same result can be achieved by omitting rectification, which we used here for greater computational efficiency.


\paragraph{DoG convolution.} Spatial summation over the RF and center-surround antagonism is modeled by incorporating a set of fixed-weight DoG filters. The filters are parameterized by the center and surround radii and peak sensitivity ratio obtained from a prior neurophysiological study \cite{CRONER19957} (see Supplementary Material Section \ref{sec:implementation_details}). We simulate biological color-opponent pathways characteristic of the different types of ganglion cells  \cite{10.3389/fneur.2021.661938, Wiesel1966-zq}. Midget cells incorporate red-green, green-red, and blue-yellow opponency, whereas parasol cells reflect achromaticity by incorporating a  DoG filter with no color tuning. Thus, in total, the Retinablock comprises four parallel channels.

\begin{figure}[t!]
\centering
\begin{tikzpicture}
\pgftext{\includegraphics[width=1.0\linewidth]{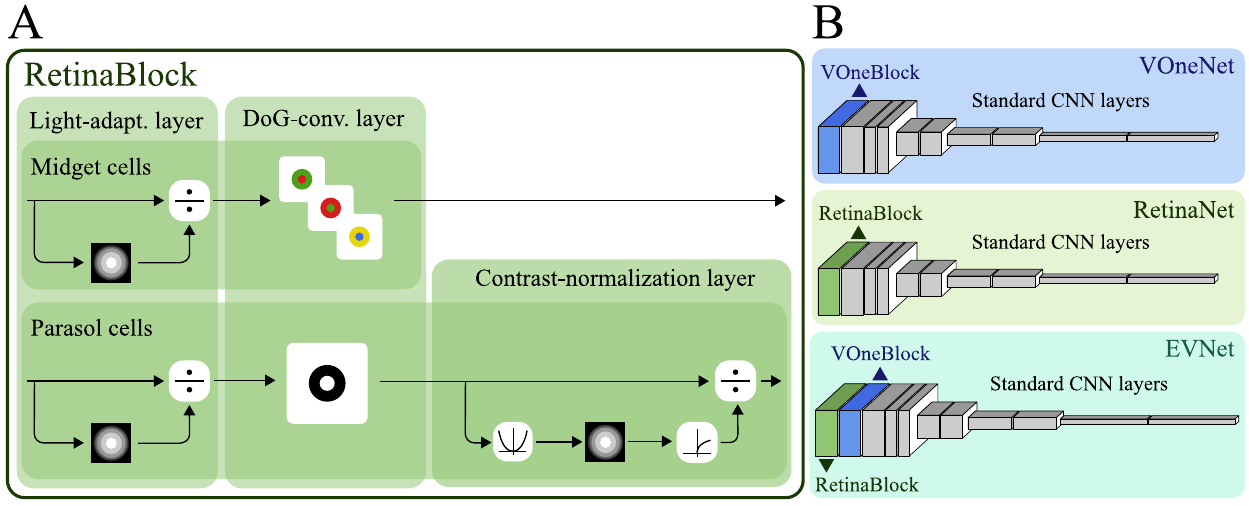}};
\end{tikzpicture}
\caption{\textbf{Simulating early visual processing of primates as CNN front-end blocks.} \textbf{A} The RetinaBlock integrates a light-adaptation layer, a DoG convolutional layer with color-opponent pathways for migdet cells, and, for parasol-cells, a contrast-normalization layer. \textbf{B} VOneNet, RetinaNet and EVNet comprise an initial block designed to simulate a specific stage of the visual system, followed by a standard CNN architecture. The VOneNet includes the VOneBlock; the RetinaNet includes the RetinaBlock; and the EVNet includes both.}\label{fig:1}
\end{figure}

\paragraph{Light adaptation.} The light-adaptation layer computes local spatial contrast, through subtractive and divisive normalization, using the local mean luminance \cite{Carandini2011-xb, MANTE2008625, berardino2018eigendistortions} (see Supplementary Material Section \ref{sec:implementation_details} for implementation details).

\paragraph{Contrast normalization.} The contrast-normalization layer divides activations by their local contrast \cite{Carandini2011-xb, MANTE2008625, berardino2018eigendistortions} (see Supplementary Material Section \ref{sec:implementation_details}), mimicking the adaptive processes observed in early vision \cite{Carandini2011-xb, Bonin2005-cu, MANTE2008625, RaghavanENEURO.0515-22.2023, Mante2005-kc}. This layer is specific to the parasol-cell channel to simulate the heightened nonlinearity of these cells in response to contrast \cite{Lee1994-co, RaghavanENEURO.0515-22.2023}.

We trained four seeds of each biologically-inspired model variant (VOneNet, RetinaNet, and EVNet), as well as the standard model, for two different CNN architectures (ResNet18 and VGG16) on Tiny ImageNet \cite{le2015tiny}. We evaluated clean accuracy on the Tiny ImageNet validation set and robustness using Tiny ImageNet-C \cite{hendrycks2019benchmarking}, which contain 75 different types of perturbations grouped in four categories (see Supplementary Material Section \ref{sec:training_details}). \footnote{The code is available at \url{https://www.github.com/lucaspiper99/retinanets-evnets}.}

\section{Results} \label{sec:results}

\subsection{The RetinaBlock simulates empirical retinal ganglion cell response properties}

We conducted a set of in-silico experiments using drifting grating stimuli, to assess single-cell response properties of the RetinaBlock and study how  VOneBlock responses are affected when coupled with the RetinaBlock. The activation of the centrally-stimulated cell was recorded and subsequent Fourier analysis was performed. We then examined how each response varied as a function of spatial frequency (SF) and contrast (see Supplementary Material Section \ref{sec:implementation_details_gratings}).

\begin{figure}[h!]
\centering
\begin{tikzpicture}
\pgftext{\includegraphics[width=1.0\linewidth]{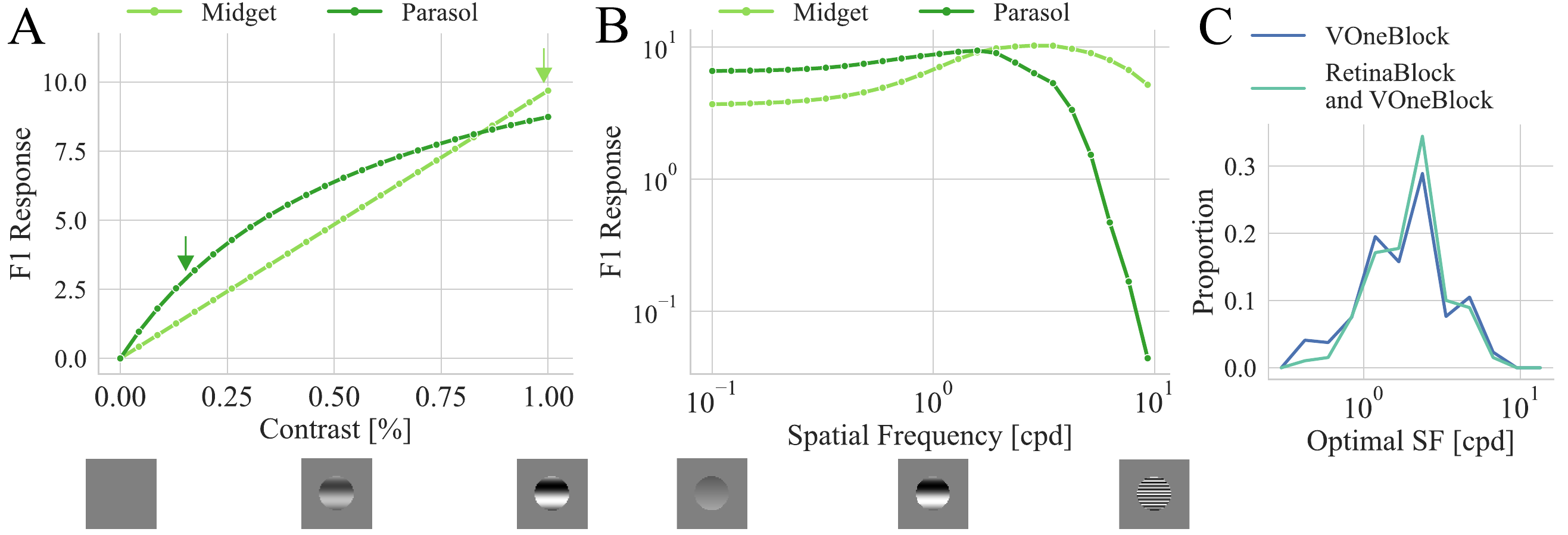}};
\end{tikzpicture}
\caption{\textbf{RetinaBlock simulates retinal response properties to SF and contrast.} \textbf{A} Contrast sensitivity curves of example midget and parasol cells of the RetinaBlock with corresponding stimuli below. Arrows denote where logarithmic saturation begins by fitting a log contrast response function \cite{RaghavanENEURO.0515-22.2023}. \textbf{B} SF tuning curves with the corresponding grating stimuli below. Activation range differs across cell types due to the compression introduced by the contrast-normalization layer. The optimal SF is 4.2 cycles per degree (cpd) for midget cells and 1.0 cpd for parasol cells. \textbf{C} Distribution of optimal SF for VOneBlock cells with and without prior RetinaBlock processing.}\label{fig:2}
\end{figure}

Midget and parasol cells are known to exhibit discriminant properties when responding to contrast changes \cite{RaghavanENEURO.0515-22.2023, Solomon2006-zt, Lee1994-co}. Parasol cells exhibit a higher initial slope and a higher degree of contrast saturation than midget cells, whereas midget cells have a higher half-response constant \cite{RaghavanENEURO.0515-22.2023}. This pattern is also observed for RetinaBlock cells (see Figure \ref{fig:2} A). Additionally, the SF response curve of RetinaBlock cells delineates a DoG spectra (see Figure \ref{fig:2} B), consonant with empirical measurements of retinal ganglion cell responses \cite{CRONER19957, LINSENMEIER19821173}. 

Interestingly, the introduction of the RetinaBlock before the VOneBlock did not greatly change the response of V1 cells in terms of their SF tuning. The mean optimal SF was slightly reduced from 2.97cpd to 2.85cpd with the introduction of the RetinaBlock (see Figure \ref{fig:2} C). In addition, the SF bandwidth \cite{doi:10.1152/jn.1976.39.6.1334} of V1 cells was also only mildly affected (data not shown).




\subsection{RetinaNets improve robustness against corruptions}

\begin{figure}[ht]
\centering
\begin{tikzpicture}
\pgftext{\includegraphics[width=1.0\linewidth]{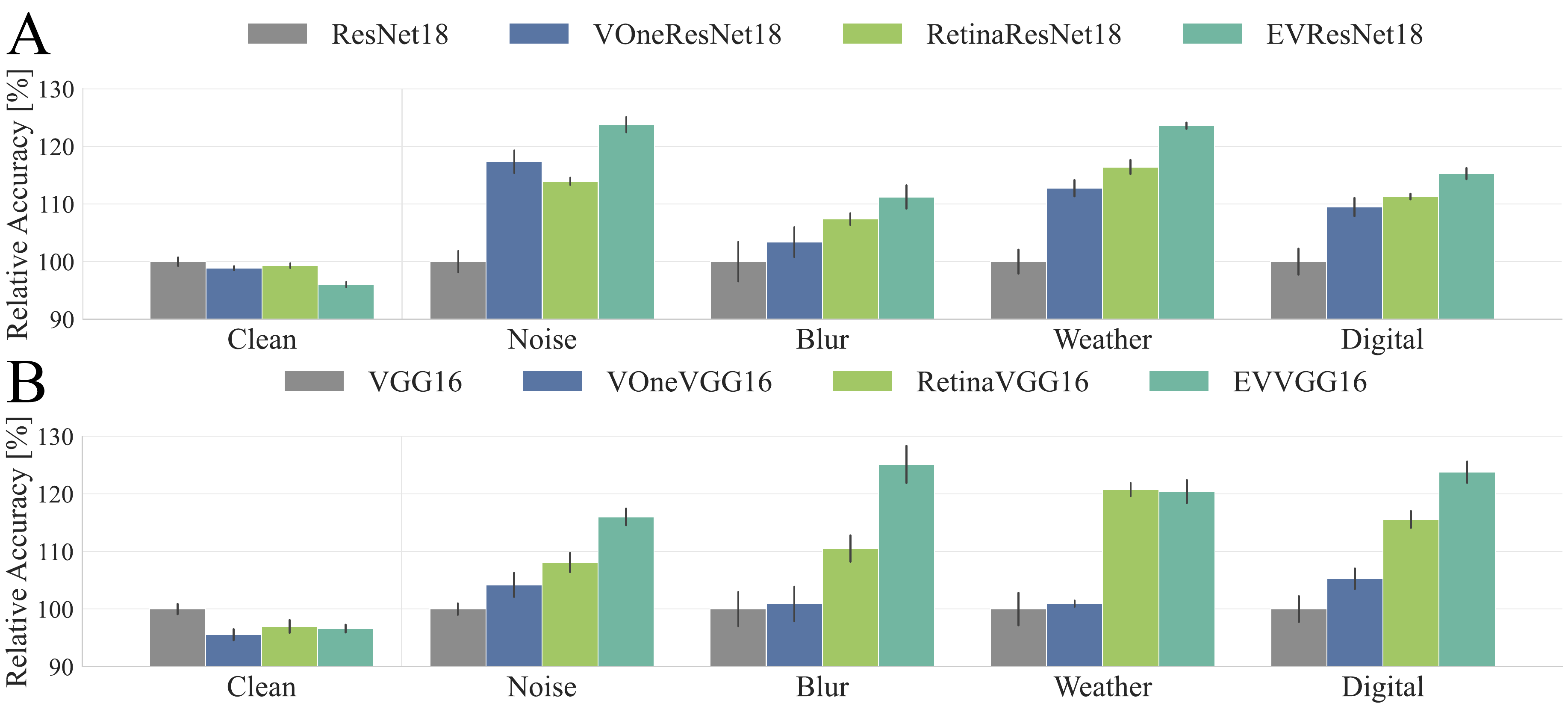}};
\end{tikzpicture}
\caption{\textbf{RetinaNets improve robustness to all corruption categories and EVNets further improve upon VOneNets and RetinaNets.} \textbf{A} Relative accuracy (normalized by ResNet18 accuracy) on clean images and all corruptions categories for the base ResNet18, VOneResNet18, RetinaResNet18 and EVResNet18 (see Table \ref{tab:abs_acc_corr} and Figure \ref{fig:A1} for absolute accuracies). Bars represent the mean and error bars represent the SE ($n$ = 4 seed initializations). \textbf{B} Relative accuracy (normalized by VGG16) on clean images and all corruptions categories for models based in the VGG16 architecture (absolute accuracies in Table \ref{tab:abs_acc_vgg16}).}\label{fig:3}
\end{figure}

Similarly to the VOneNets \cite{Dapello2020.06.16.154542, baidya2021combining}, we observed a small drop in clean accuracy for RetinaNets when compared to the base model (relative accuracy of 99.3\% for the RetinaResNet18 and 97.0\% for RetinaVGG16, see Figure \ref{fig:3}). RetinaNets exhibit a slightly better performance relative to VOneNets.

In terms of robustness, RetinaNets show an improvement in mean accuracy on the Tiny ImageNet-C dataset (see Figure \ref{fig:3}) when compared to both the base models and the VOneNet variants. RetinaResNet18 achieves an overall relative gain of 12.7\%  (10.3\% for VOneResNet18). RetinaResNet18 consistently improved robustness across all corruptions (see Table \ref{tab:abs_acc_corr} for absolute accuracies). Similar results were observed for the VGG16-based models with RetinaVGG16 improving 13.7\% against only 2.8\% of the VOneVGG16 (see Table \ref{tab:abs_acc_vgg16}).


\subsection{The RetinaBlock-VOneBlock interaction provides cumulative robustness gains}

Like before, EVNets, which incorporate both a RetinaBlock and a VOneBlock, also show a small decline in accuracy, when compared to the base models (96.0\% relative accuracy for EVResNet18 and 96.6\% for EVVGG16) and barely underperform the corresponding RetinaNet models.  

Interestingly, EVNets improves robustness across all corruption categories, independently of the back-end architecture, with an overall relative gain of 18.1\% for EVResNet18 and 20.7\% for EVVGG16. When using the same base model, EVNets consistently outperform VOneNets and outperform RetinaNets in most, except a few specific corruptions types (see Tables \ref{tab:abs_acc_corr} and  \ref{tab:abs_acc_vgg16} and Figure \ref{fig:3}).

\section{Discussion} \label{sec:discussion}




A primary objective in computer vision is the development of models that exhibit enhanced robustness to images under distribution shifts. This is crucial if one wishes to deploy these models in critical real-world applications. In this study, we introduced two novel CNN families that demonstrated improved model robustness across many types of image corruptions while maintaining relatively high clean accuracy. Though akin to prior DoG filtering approaches \cite{10.1007/978-3-031-44204-9_33, EVANS202296} and traditional normalization methods \cite{NIPS2012_c399862d, 5459469}, the RetinaBlock enforces luminance and contrast normalization coupled with the band-pass behaviour of the DoG filters, enhancing feature selectivity. Additionally, the RetinaBlock achieves this by incorporating a set of mechanisms that consistently follow biologically plausibility with no additional training overhead. Furthermore, while these models do not fully resolve the challenge of robust generalization, our findings indicate that progress can be made by integrating biologically-plausible models of the primate visual system into deep learning architectures. The results from this study demonstrate that simulating early visual processing as multi-stage front-ends can enhance CNN robustness to image corruptions with minor trade-offs. Specifically, the cumulative VOneBlock and RetinaBlock gains indicate that these blocks contribute to different types of invariance, yielding stacked gains in model robustness. In fact, the RetinaBlock focuses on luminance and contrast invariance, the VOneBlock focuses more on spatial and polarity invariance.

Although these improvements are noteworthy, they are not without limitations. Improvements in robustness are consistently paired with a slight decrease in accuracy on clean images. Furthermore, the relative robustness gains are not consistent across base models. For instance, blur is the corruption type in which EVResNet18 performs the worst, and, simultaneously, the one in which EVVGG16 performs the best. Moreover, the slightly lower gains observed with the VOneVGG16 model, compared to previous implementations of VOneNets \cite{Dapello2020.06.16.154542, baidya2021combining}, suggest that the choice of back-end architecture may play an important role in determining the effectiveness of these neural front-end enhancements. Investigating an alternative back-end architecture or integration that better synergizes with these front-end blocks can potentially unlock further improvements in robustness. Besides this, future research may explore alternative directions. For example, assess how these robustness gains scale to larger input images and to different out-of-domain (OOD) datasets. Likewise, studying the individual contributions of each component within the RetinaBlock can potentially elucidate their relative importance in improving model robustness. Furthermore, exploring how different color contributions shape model performance could provide insights into optimizing color processing in artificial vision systems. Finally, future work may also explore the introduction of independent neural noise in the different neurobiological stages to potentially shape a more robust network while mimicking the inherent variability in primate visual systems.


\clearpage

\section*{Acknowledgements}

This work was supported by the project Center for Responsible AI reference no. C628696807-00454142, financed by the Recovery and Resilience Facility, by project PRELUNA, grant PIDC/CCIINF/4703/2021, and by Fundação para a Ciência e Tecnologia, under grant UIDB/50021/2020.

\printbibliography


\newpage

\appendix 
\appendixpage

\section{Datasets} \label{sec:datasets}

\subsection{Tiny ImageNet} \label{sec:datasets_tiny}

We used the Tiny ImageNet dataset for model training and evaluating model clean accuracy \cite{le2015tiny}. Tiny ImageNet contains 100.000 images of 200 classes (500 for each class) downsized to 64$\times$64 colored images. Each class has 500 training images, 50 validation images and 50 test images. Tiny ImageNet is publicly available at \url{http://cs231n.stanford.edu/tiny-imagenet-200.zip}.

\subsection{Tiny ImageNet-C (common corruptions)} \label{sec:datasets_tinyc}

For evaluating model robustness to common corruptions we used the Tiny ImageNet-C dataset  \cite{hendrycks2019benchmarking}. The Tiny ImageNet-C dataset consists of 15 different corruption types, each at 5 levels of severity for a total of 75 different perturbations, applied to validation images of Tiny ImageNet. The individual corruption types are: Gaussian noise, shot noise, impulse noise, defocus blur, glass blur, motion blur, zoom blur, snow, frost, fog, brightness, contrast, elastic transform, pixelate and JPEG compression. The individual corruption types are grouped into 4 categories: noise, blur, weather, and digital effects. The Tiny ImageNet-C is publicly available at \url{https://github.com/hendrycks/robustness} under Creative Commons Attribution 4.0 International.

\section{Models} \label{sec:models}

\subsection{VOneNets} \label{sec:vonenets}

\paragraph{VOneNet model family.} VOneNets \cite{Dapello2020.06.16.154542} are CNNs with a biologically-constrained fixed-weight front-end that simulates V1, the VOneBlock -- a linear-nonlinear-Poisson (LNP) model of V1 \cite{RUST2005945}, consisting of a fixed-weight Gabor filter bank (GFB) \cite{Jones1987-dy}, simple and complex cell \cite{Adelson:85} nonlinearities, and neuronal stochasticity \cite{Softky1993-te}. The GFB parameters are generated by randomly sampling from empirically observed distributions of preferred orientation, peak SF, and size of RFs \cite{DEVALOIS1982545, DEVALOIS1982531, ringach2002spatial_dtructure}, the channels are divided equally between simple- and complex-cells (256 each), and a Poisson-like stochasticity generator is used. Code for the VOneNet model family is publicly available at \url{https://github.com/dicarlolab/vonenet} under GNU General Public License v3.0.

\paragraph{Adapting the VOneBlock to Tiny ImageNet.} Due to the difference in input size in comparison to ImageNet, we set the stride of the GFB at two instead of four such that the output of the VOneBlock does not have a very small spatial map and adjusted the input field of view to 2deg for Tiny ImageNet instead of 8deg for ImageNet to account for the fact that images in the first represent a narrower portion of the visual space \cite{baidya2021combining}. Given the new resolution, we bounded the SF of the GFB between 0.5cpd and 11.3 cpd. We also removed the stochasticity generator, so that the models are noise-free. We set the GFB to uniformly sample a single channel from the input to be processed by the VOneBlock, regardless of prior processing by the RetinaBlock.

\subsection{ResNet18-based models} \label{sec:models_resnet18}

The VOneResNet18 and EVResNet18 were built by replacing the first block (convolution, normalization, non-linearity and pooling layers) of ResNet18 \cite{he2015deep} by the VOneBlock, a trainable bottleneck layer and, for EVNets, the RetinaBlock. For these models along with the base ResNet18, we used a modified version of the Torchvision ResNet18 model \cite{he2015deep}. In the original ResNet18, the first block has a combined stride of four (two from the convolution layer and two from the maxpool layer), which is replaced by VOneBlock in VOneResNet18 and in EVResNet18. To all models comparable, we adjusted ResNet18 to have a stride of one in the first convolutional layer and two in the maxpool layer, resulting in a combined stride of two, similar to VOneBlock. This ResNet18 variant (58.93\% accuracy) outperformed the standard version (50.45\% accuracy) on clean Tiny ImageNet images after identical training.

\subsection{VGG16-based models} \label{sec:models_vgg16}

The VOneVGG16 and EVVGG16 were created by replacing the first convolution and non-linearity of VGG16 \cite{simonyan2015deep} with the VOneBlock, a trainable bottleneck layer, and for EVNets, the RetinaBlock. To ensure comparability among models without excessively small feature maps, we modified the first layer of the base model to have a stride of two, kernel size of seven, and padding of three. Additionally, following a prior adaptation for Tiny ImageNet \cite{wu2017tiny}, we reduced the intermediate fully-connected layer size from 4096 to 2048 and removed the last max-pooling layer.

\subsection{Training details} \label{sec:training_details}

Each model was trained on one of the following two configurations: (i) 32GB NVIDIA V100 GPU with Python 3.9.7 and PyTorch 2.0.0+cu117; or (ii) 48GB NVIDIA A40 GPU with Python 3.10.13 and PyTorch 2.2.0+cu118; taking, on average, 75min to train and 15min to test.

\paragraph{Preprocessing.} During training, preprocessing included scaling images by a factor randomly sampled between 1 to 1.2, randomly rotating images by an angle between -30 to 30 degrees, random vertical/horizontal shifting between -5\% to 5\% of the image width/height, and horizontal flipping with a random probability of 0.5. Images were also normalized by subtraction and division by [0.5, 0.5, 0.5], for models that did not include the RetinaBlock (base models and VOneNets). During evaluation, preprocessing only involved image normalization, for models with no RetinaBlock.

\paragraph{Loss function and optimization.} The loss function was given by a cross-entropy loss between image labels and model predictions (logits). For optimization, we used Stochastic Gradient Descent with momentum 0.9 and a weight decay 0.0005. The learning rate was divided by 10 whenever there is no improvement in validation loss for 5 consecutive epochs. All models were trained using a batch size of 128 images. Models based on the ResNet18 architecture trained for 60 epochs with an initial learning rate of 0.1, whereas VGG16-based models trained for 100 epochs with an intial learning rate of 0.01.

\section{Implementation details} \label{sec:implementation_details}

\subsection{RetinaBlock parameterization} \label{sec:implementation_details_retinablock}

To preserve dimensionality and gain across activation maps, we used filters with unity sum, convolutions with unity stride and reflective padding with size equal to the kernel size integer division by two.

\paragraph{DoG convolution.} The filters are parameterized by the center and surround radii and corresponding peak contrast sensitivities from the distribution medians reported by Croner and Kaplan \cite{CRONER19957} (Table 1, under P-cells within the eccentricity range of 0-5 degrees and under M-cells within the range of 0-10). The kernel size of both cell types was defined as to be capable of producing up to 95\% of the integrated response of the surround, yielding 21px for midget cells and 65px for parasol cells.

\paragraph{Light adaptation.} The light-adaptation subtracts and divides the input by the local mean luminance as formulated in Equation \ref{eq:la}, where $\mathbf{x}$ denotes the input image and $\mathbf{x_{LA}}$ is the output of the light-adaptation layer. The mean luminance is computed by convolving the input with a Gaussian filter, $\mathbf{w_{LA}}$, originating a single channel (average RGB luminance). The kernel and the Gaussian width are set to be four times that of the surround Gaussian in the midget DoG kernel (2.625deg and 85px, respectively). This pooling size was chosen to provide a localized luminance estimation, without introducing a low-SF cut in the cell's SF tuning curve nor increasing optimal SF past 3 cpd \cite{Alitto2008-df, Levitt2001-ar}.

\begin{equation} 
    \mathbf{x_{LA}} = \frac{\mathbf{x} - \mathbf{x} * \mathbf{w_{LA}}}{\mathbf{x} * \mathbf{w_{LA}}}
    \label{eq:la}
\end{equation}

\paragraph{Contrast normalization.} The computation of the local contrast in the contrast-normalization layer is described by Equation \ref{eq:cn}. $\mathbf{x_{DoG}}$ is the activations from the DoG-convolutional layer and $\mathbf{x_{CN}}$ are the activations from the contrast-normalization laey. Parameters follow prior empirical studies on mammalian LGN: the half-response contrast $c_{50}$ is set to 0.3 \cite{Bonin2005-cu, RaghavanENEURO.0515-22.2023} and the weights of the suppressive field $\mathbf{w_{CN}}$ describe a Gaussian kernel coextensive with the the cell's surround \cite{MANTE2008625, Bonin2005-cu} (0.72deg radius Gaussian and 65px kernel).

\begin{equation}
    \mathbf{x_{CN}} = \frac{\mathbf{x_{DoG}}}{c_{50} + \sqrt{\mathbf{x_{DoG}}^2 * \mathbf{w_{CN}}}}
    \label{eq:cn}
\end{equation}

\subsection{Drifting grating stimuli} \label{sec:implementation_details_gratings}

We presented 12 frames of drifting sine-wave gratings with phase shifts of 30 degrees in the interval [0, 360[ degrees. Grating orientation was set to horizontal and the diameter of the gratings was set to 1 degree of the field of vision. The background area not covered by the grating was set to 50\% gray. For the SF tuning curve, SFs ranged logaritmically from 0.1 cpd to 9.3 cpd (2 cpd below Nyquist SF) in 24 steps. For contrast sensitivity, grating contrast (defined as in Equation \ref{eq:contrast}, where $L_{min}$ and $L_{max}$ denote the minimum and the maximum grating luminance) varied linearly from 0 to 1 in 24 steps. For complex cells of the VOneBlock, we extracted the mean response (F0), whereas for all remaining cells, we extracted the first harmonic amplitude (F1) \cite{DEVALOIS1982545}.

\begin{equation}
    C=\frac{L_{max}-L_{min}}{L_{max}+L_{min}}
    \label{eq:contrast}
\end{equation}

\section{Detailed Results} \label{sec_detailed_results}

\begin{table}[h!]
\centering
\caption{\textbf{Absolute top-1 accuracies of ResNet18, VOneResNet18 and
RetinaResNet18 and EVResNet18.} Clean images and 15 types of common image corruptions (averaged over five perturbation severities). The value in parenthesis represents the standard error of the mean ($n$ = 4 seeds).}
\label{tab:abs_acc_corr}
\begin{tabular}{@{}lcccccccc@{}}
\toprule
 &  & \multicolumn{3}{c}{Noise} & \multicolumn{4}{c}{Blur} \\ \cmidrule(r{10pt}){3-5} \cmidrule(r){6-9}
 & Clean & Gaussian & Shot & Impulse & Defocus & Glass & Motion & Zoom \\
Model & {[}\%{]} & {[}\%{]} & {[}\%{]} & {[}\%{]} & {[}\%{]} & {[}\%{]} & {[}\%{]} & {[}\%{]} \\ \midrule
\multirow{2}{*}{ResNet18} & \textbf{58.00} & 19.70 & 22.98 & 21.90 & 14.02 & 18.80 & 19.01 & 15.86 \\
 & (0.43) & (0.54) & (0.46) & (0.22) & (0.58) & (0.49) & (0.65) & (0.61) \\
\multirow{2}{*}{VOneResNet18} & 57.35 & 23.84 & 27.81 & 24.13 & 14.48 & 19.24 & 20.01 & 16.26 \\
 & (0.20) & (0.52) & (0.55) & (0.23) & (0.38) & (0.26) & (0.64) & (0.49) \\
\multirow{2}{*}{RetinaResNet18} & 57.60 & 23.24 & 27.19 & 23.15 & 15.14 & 19.53 & 20.84 & 17.17 \\
 & (0.24) & (0.12) & (0.21) & (0.09) & (0.18) & (0.06) & (0.28) & (0.22) \\
\multirow{2}{*}{EVResNet18} & 55.70 & \textbf{25.42} & \textbf{30.27} & \textbf{24.24} & \textbf{16.02} & \textbf{19.95} & \textbf{21.54} & \textbf{17.78} \\
 & (0.27) & (0.35) & (0.37) & (0.23) & (0.40) & (0.21) & (0.33) & (0.47) \\ \bottomrule
\end{tabular}
\end{table}

\begin{table}[h!]
\centering
\begin{tabular}{@{}lcccccccc@{}}
\toprule
 & \multicolumn{4}{c}{Weather} & \multicolumn{4}{c}{Digital} \\ \cmidrule(r{10pt}){2-5} \cmidrule(r){6-9}
 & Snow & Frost & Fog & Bright. & Contrast & Elastic & Pixelate & JPEG \\
Model & {[}\%{]} & {[}\%{]} & {[}\%{]} & {[}\%{]} & {[}\%{]} & {[}\%{]} & {[}\%{]} & {[}\%{]} \\ \midrule
\multirow{2}{*}{ResNet18} & 23.74 & 24.81 & 20.66 & 9.30 & 24.07 & 37.44 & 31.26 & 26.89 \\
 & (0.44) & (0.40) & (0.67) & (0.26) & (0.85) & (0.64) & (0.86) & (0.35) \\
\multirow{2}{*}{VOneResNet18} & 27.20 & 27.59 & 23.25 & 10.05 & 27.29 & 37.55 & 36.90 & 29.37 \\
 & (0.36) & (0.35) & (0.30) & (0.19) & (0.56) & (0.26) & (0.59) & (0.51) \\
\multirow{2}{*}{RetinaResNet18} & 27.71 & 28.62 & 24.26 & \textbf{11.50} & 26.89 & \textbf{38.77} & 35.30 & 31.06 \\
 & (0.27) & (0.30) & (0.33) & (0.19) & (0.20) & (0.14) & (0.09) & (0.21) \\
\multirow{2}{*}{EVResNet18} & \textbf{30.10} & \textbf{30.18} & \textbf{25.26} & 10.87 & \textbf{29.54} & 37.61 & \textbf{38.82} & \textbf{31.82} \\
 & (0.18) & (0.16) & (0.15) & (0.12) & (0.55) & (0.43) & (0.29) & (0.35) \\ \bottomrule
\end{tabular}
\end{table}

\begin{table}[h!]
\centering
\caption{\textbf{Absolute top-1 accuracies of VGG16, VOneVGG16 and RetinaVGG16 and EVVGG16.} Clean images and 15 types of common image corruptions (averaged over five perturbation severities). The value in parenthesis represents the standard error of the mean ($n$ = 4 seeds).}
\label{tab:abs_acc_vgg16}
\begin{tabular}{lcccccccc}
\toprule
 &  & \multicolumn{3}{c}{Noise} & \multicolumn{4}{c}{Blur} \\ \cmidrule(r{10pt}){3-5} \cmidrule(r){6-9} 
 & Clean & Gaussian & Shot & Impulse & Defocus & Glass & Motion & Zoom \\
Model & {[}\%{]} & {[}\%{]} & {[}\%{]} & {[}\%{]} & {[}\%{]} & {[}\%{]} & {[}\%{]} & {[}\%{]} \\ \midrule
\multirow{2}{*}{VGG16} & \textbf{45.10} & 18.58 & 21.46 & 18.86 & 10.60 & 13.71 & 13.89 & 11.80 \\
 & (0.40) & (0.19) & (0.29) & (0.15) & (0.35) & (0.29) & (0.44) & (0.42) \\
\multirow{2}{*}{VOneVGG16} & 43.10 & 19.60 & 22.21 & 19.56 & 10.70 & 13.45 & 14.39 & 11.90 \\
 & (0.43) & (0.53) & (0.52) & (0.19) & (0.40) & (0.30) & (0.44) & (0.39) \\
\multirow{2}{*}{RetinaVGG16} & 43.74 & 20.44 & 23.93 & 19.29 & 11.83 & 14.51 & 15.80 & 13.10 \\
 & (0.50) & (0.37) & (0.29) & (0.32) & (0.32) & (0.20) & (0.36) & (0.28) \\
\multirow{2}{*}{EVVGG16} & 43.57 & \textbf{22.12} & \textbf{25.36} & \textbf{20.85} & \textbf{13.98} & \textbf{15.49} & \textbf{17.69} & \textbf{15.39} \\
 & (0.30) & (0.30) & (0.32) & (0.23) & (0.40) & (0.26) & (0.49) & (0.47) \\ \bottomrule
\end{tabular}
\end{table}

\begin{table}[h!]
\centering
\begin{tabular}{lcccccccc}
\toprule
 & \multicolumn{4}{c}{Weather} & \multicolumn{4}{c}{Digital} \\ \cmidrule(r{10pt}){2-5} \cmidrule(r){6-9}
 & Snow & Frost & Fog & Bright. & Contrast & Elastic & Pixelate & JPEG \\
Model & {[}\%{]} & {[}\%{]} & {[}\%{]} & {[}\%{]} & {[}\%{]} & {[}\%{]} & {[}\%{]} & {[}\%{]} \\ \midrule
\multirow{2}{*}{VGG16} & 18.29 & 18.35 & 14.02 & 5.28 & 17.12 & 26.92 & 23.84 & 18.57 \\
 & (0.57) & (0.54) & (0.33) & (0.05) & (0.62) & (0.46) & (0.66) & (0.35) \\
\multirow{2}{*}{VOneVGG16} & 19.67 & 17.53 & 13.94 & 4.48 & 19.81 & 26.06 & 27.05 & 19.19 \\
 & (0.15) & (0.17) & (0.12) & (0.18) & (0.55) & (0.22) & (0.49) & (0.56) \\
\multirow{2}{*}{RetinaVGG16} & 21.92 & \textbf{21.86} & \textbf{17.40} & \textbf{6.61} & 21.60 & 27.25 & 29.21 & 21.34 \\
 & (0.14) & (0.21) & (0.25) & (0.07) & (0.42) & (0.27) & (0.43) & (0.29) \\
\multirow{2}{*}{EVVGG16} & \textbf{22.66} & 21.07 & 17.27 & 6.40 & \textbf{24.34} & \textbf{28.61} & \textbf{31.23} & \textbf{22.94} \\
 & (0.45) & (0.30) & (0.27) & (0.20) & (0.51) & (0.44) & (0.37) & (0.28) \\ \bottomrule
\end{tabular}
\end{table}

\begin{figure}[h!]
\centering
\begin{tikzpicture}
\pgftext{\includegraphics[width=1.0\linewidth]{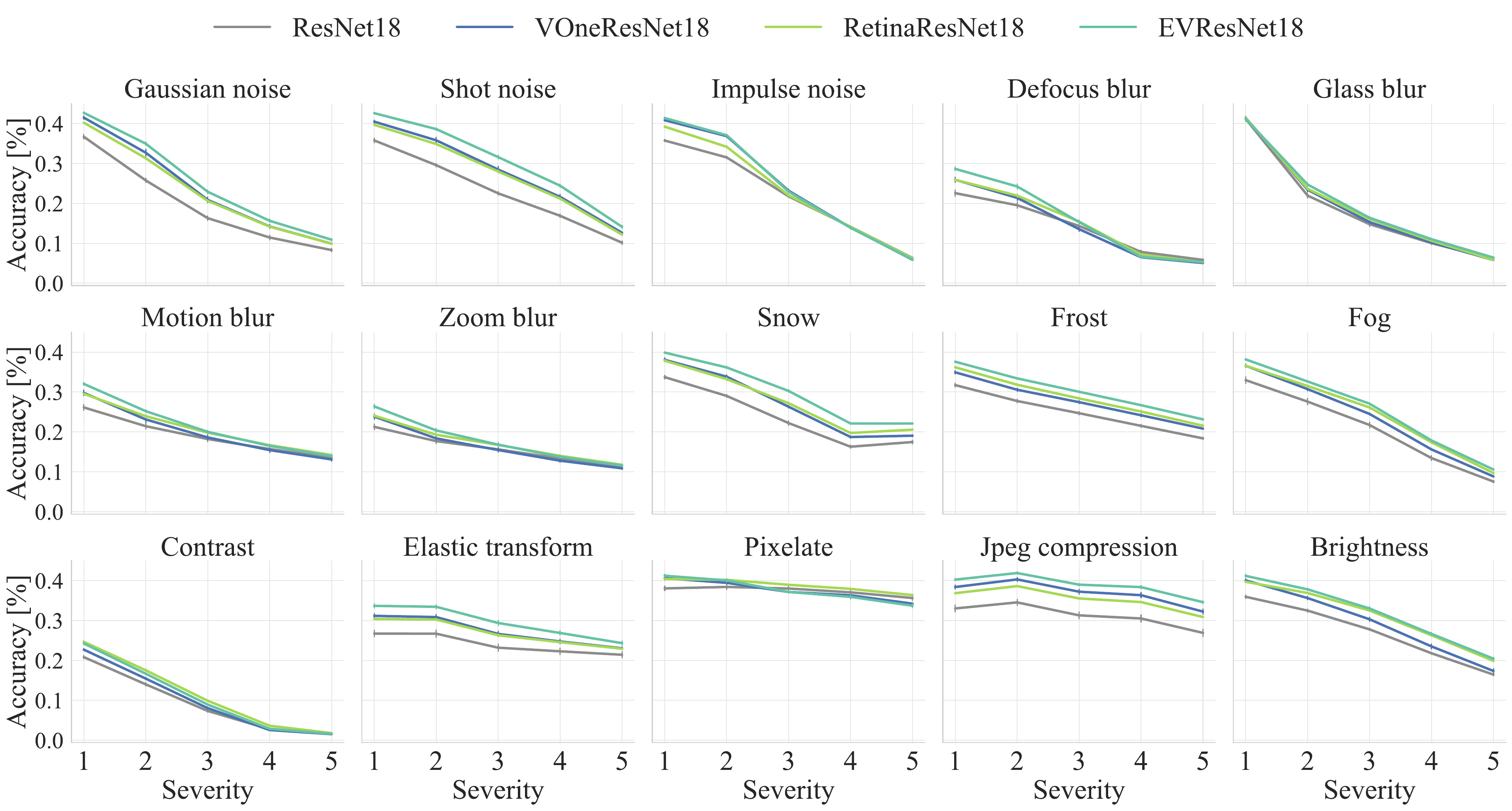}};
\end{tikzpicture}
\caption{\textbf{Absolute top-1 accuracies of ResNet18, VOneResNet18, RetinaResNet18 and EVResNet18 for 15 corruption types at 5 perturbation severity levels.} Lines represent the mean and error bars represent the standard error of the mean ($n$ = 4)}\label{fig:A1}
\end{figure}

\end{document}